\documentclass{article}
\usepackage{amsmath,graphicx,mlspconf}
\usepackage{amssymb}
\usepackage{caption}
\usepackage{subcaption}
\usepackage{amsthm}
\usepackage{color}
\usepackage{bbm}
\newtheorem{theorem}{Theorem}
\DeclareMathOperator{\supp}{supp}

\copyrightnotice{979-8-3503-2411-2/23/\$31.00 {\copyright}2023 IEEE}

\toappear{2023 IEEE International Workshop on Machine Learning for Signal Processing, Sept.\ 17--20, 2023, Rome, Italy}

\title{Uncertainty quantification for learned ISTA}
\name{Frederik Hoppe$^1$ \ Claudio Mayrink Verdun$^{2,3}$ \ Hannah Laus$^{1,2,3}$ \ Felix Krahmer$^{2,3,4}$ \ Holger Rauhut$^1$
\thanks{The first two authors contributed equally to this work. This work was supported by German Federal Ministry of Education and Research through the grant  "SparseMRI3D+ (FZK 05M20WOA)".}}
\address{$^1$Chair of Mathematics of Information Processing, RWTH Aachen University, Aachen, Germany\\
$^{2}$Department of Mathematics, Technical University of Munich, Munich, Germany\\
$^{3}$Munich Center for Machine Learning, Munich, Germany\\
$^{4}$Munich Data Science Institute, Technical University of Munich, Munich, Germany\\
\\
To appear in the proceedings of IEEE MLSP, 2023\\
(33rd IEEE International Workshop on Machine Learning for Signal Processing)
}

\begin{document}

\maketitle

\begin{abstract}
Model-based deep learning solutions to inverse problems have attracted increasing attention in recent years as they bridge state-of-the-art numerical performance with interpretability.  In addition, the incorporated prior domain knowledge can make the training more efficient as the smaller number of parameters allows the training step to be executed with smaller datasets. Algorithm unrolling schemes stand out among these model-based learning techniques. Despite their rapid advancement and their close connection to traditional high-dimensional statistical methods, they lack certainty estimates and a theory for uncertainty quantification is still elusive. This work provides a step towards closing this gap proposing a rigorous way to obtain confidence intervals for the LISTA estimator.

\end{abstract}

\section{Introduction}
In light of the recent developments in deep learning (DL), inspired by its success for computer vision \cite{lecun2015deep}, also the computational view on inverse problems has significantly changed over the last years \cite{arridge2019solving}. Now most state-of-the-art approaches involve some type of training step where the reconstruction method is adapted to a ground truth data set and their corresponding measurements via supervised learning techniques.

While fully trained DL approaches are known to exhibit excellent performance, they also pose some challenges, especially for medical applications. Namely, due to their black-box nature, it is very difficult to quantify the quality of their reconstruction, which is crucial for certification processes and widespread use. Moreover, the lack of interpretability of high-performing architectures due to their black-box nature and the need to large training datasets makes the development of any theory a very challenging task \cite{grohs2022mathematical}. This is why it has been an active field of study to connect these approaches to model-based methods, as they are interpretable and easier to analyze in mathematical terms. In particular, parsimonious models like sparsity are much better understood for applications and have hence been proposed as inspiration for synergetic approaches in a number of works.
On the one hand, such models have been demonstrated to be quite expressive allowing to capture the nature of various kinds of data, both for statistical problems and for real-world signal and image processing applications.
On the other hand, pioneered by convex methods such as the LASSO \cite{tibshirani1996regression} or Basis Pursuit \cite{chen2001atomic}, very efficient and scalable recovery methods with reconstruction guarantees have been developed for many such models \cite{Foucart.2013}, making them feasible for practical problem sizes. Consequently, they are widely used in real-world applications such as magnetic resonance imaging (MRI) \cite{lustig2008compressed}.

A very popular way to link this viewpoint to neural network (NN) approaches is based on the observation that many iterative algorithms can be expressed in a recurrent neural network form, where each layer represents an iteration; often referred as unrolling \cite{monga2021algorithm}. This yields a NN method with recovery guarantees when the weights are not trained, but taken directly from the iterative algorithm. 
The idea is then that optimizing the weights, i.e., incorporating a training step, should make the performance better and hence ideally inherit the theoretical guarantees of traditional compressive sensing approaches while at the same time exhibiting a predictive power comparable to sophisticated large black-box schemes. 

This line of research started with the seminal paper \cite{gregor2010learning} that proposed the Learned Iterative Shrinkage Thresholding Algorithm (LISTA), a data-driven approach where each layer of the network represents an iteration of the (now) classical ISTA algorithm \cite{daubechies2004iterative}. The design of the layers aims to emulate the free parameters of each iteration in a trainable way. In particular, such methods produce a better solution with
significantly fewer layers/iterations as compared to the original ISTA. Several theoretical and practical developments followed since the original paper \cite{liu2019alista,chen2018theoretical,zheng2022hybrid,ablin2019learning}.

Despite several recent advances in theory and applications of unrolled algorithms \cite{scarlett2022theoretical, shlezinger2022model}, if the signal of interest is corrupted by noise, it is hard to assess the quality of the estimator for the underlying signal. In particular, in critical applications like medical imaging, it is essential to guarantee a high level of exactness in recovery and to perform uncertainty quantification (UQ). In the high-dimensional context, this can be particularly challenging and most of the algorithms do not come with UQ results such as confidence intervals. A prominent exception in the classical optimization theory is the debiased LASSO \cite{Zhang.2014, vandeGeer.2014, Javanmard.2018} where the LASSO estimator is debiased by applying an additional correction term and thus shown to be asymptotically normal. This asymptotic normality allows for the construction of confidence intervals. The debiased LASSO is extended from the LASSO formulation, involving the $\ell_1$-norm, to estimators that are regularized by a convex function \cite{10.1214/22-AOS2243}. To the best of the author's knowledge, there have been no approaches for debiasing an estimator that is given by a data-driven procedure such as the LISTA.

\textbf{Our contribution:}
In this paper, we formulate the debiased LISTA which is, to the best of the author's knowledge, the first UQ methodology with guarantees for confidence intervals for this type of data-driven estimator. In Section \ref{sec:background} we summarize the existing theory of sparse recovery, the learned ISTA algorithm, and the debiased LASSO. Section \ref{sec:gerneral_estimators} generalizes the debiased theory to LISTA. Then, in Section \ref{sec:conf_regions}, we provide uncertainty quantification by constructing confidence intervals for the underlying signal with the learned estimator. In Section \ref{sec:numerics} we conclude with numerical experiments that illustrate the accuracy of the method.

\section{Background and related work}\label{sec:background}
In this section, we provide a summary of the existing theory for sparse recovery, LISTA, and the debiased LASSO. We derive an extended theory for constructing a debiased LISTA estimator from this theory.

\subsection{Sparse recovery}
Let $A \in \mathbb{R}^{m \times N}$ be a dictionary with atoms $ a_1^T, \dots, a_m^T$ and $ b = (b_1, \dots b_m) \in \mathbb{R}^{m} $ a data vector. We consider the high dim. regression model with additive white Gaussian noise
\begin{equation}\label{Cmodel}
b=A x^*+\varepsilon, \qquad N\gg m,
\end{equation}
where $x^*\in\mathbb{R}^N$ is $s_0$-sparse and $\varepsilon\sim\mathcal{N}(0,\sigma^2 I_{m\times m})$ is the noise vector with independent components $\varepsilon_i$.

A well-studied estimator is the Least Absolute Shrinkage and Selection Operator (LASSO) \cite{tibshirani1996regression}, denoted by $\hat{x}$, that is given by the minimizer of the problem
\begin{equation}\label{eq:LASSO}
    \min\limits_{x\in\mathbb{R}^N}\frac{1}{2m}\Vert Ax-b\Vert_2^2+\lambda\Vert x\Vert_1,
\end{equation}
where the parameter $\lambda=\lambda(N,m,\sigma) \in \mathbb{R}$ balances the data fidelity term and sparsity induced by the $\ell_1$-norm. One prominent algorithm to solve the optimization problem \eqref{eq:LASSO} is the Iterative Shrinkage Thresholding Algorithm (ISTA) \cite{ISTAfirst} which iterates for $k=1,2,\dots,K$
\begin{equation}\label{eq:ista}
    x^{k+1}=\mathcal{S}_{\lambda}\left((I_{N\times N}-\frac{1}{\mu}A^TA)x^k+\frac{1}{\mu}A^Tb\right)
\end{equation}
with a stepsize parameter $\mu>0$ and the soft-thresholding operator
$\mathcal{S}_{\lambda}(x)=\operatorname{sgn}(x)\cdot\max(\vert x\vert-\lambda,0).$ Defining $W_1= \frac{1}{\mu}A^T$, known as the filter matrix, and $W_2=I_{N\times N}-\frac{1}{\mu}A^TA$, known as the mutual inhibition matrix, \eqref{eq:ista} reduces to
\begin{equation}\label{eq:lista}
    x^{k+1}=\mathcal{S}_{\lambda}\left(W_2x^k+W_1b\right).
\end{equation}

\subsection{Learned ISTA}

The seminal paper \cite{gregor2010learning} interpreted each of these iterations as a single layer of a recurrent neural network (RNN) and explored the network $x^{k+1}=\mathcal{S}_{\lambda}\left(W_2^kx^k+W_1^kb\right)$ with learnable parameters $W_1^k$ and $W_2^k$. They noted that this procedure with a (small) fixed number of layers ($k$ usually is around 10 to 20) can learn a very precise solution compared to the required number of iterations for ISTA to converge. Later, the work \cite{chen2018theoretical} provided convergence guarantees by assuming an asymptotic coupling between the weight matrices $W_1^k$ and $W_2^k$, i.e.,  $W_2^k=I_{N\times N}-W_1^kA$. Therefore the network can be written as $x^{k+1}=S_{\lambda^k}(x^k+(W^k)^T(b-Ax^k))$ with free trainable parameters $(W^k,\lambda^k)_{k=1}^K$. This became known as LISTA-CP, which reduces the number of trainable parameters without degrading the predictive power of the network. 
To train the network we utilize the training dataset $\{x_i^0, b_i \}_{i=1}^n$, which is sampled from some distribution $(x^0,b) \sim \mathcal{D}$. The parameters $W=\{W^k\}_{k=1}^K$ and $\lambda=\{\lambda^k\}_{k=1}^K$ are subject to learning for the number of iterations $K$. During the training process the loss
 $   \min_{\lambda, W}\frac{1}{n} \sum_{i=1}^n  \lVert x_i^k(\lambda, W, b_i, x_i^0)-x_i^* \rVert_2^2$,
is minimized by SGD-type algorithms.

\subsection{Debiased LASSO}
The works \cite{Javanmard.2014, Javanmard.2018, vandeGeer.2014} initiated the theory for the debiasing procedure for the LASSO. Most contributions about this technique assume that the measurement matrix $A\in\mathbb{R}^{m\times N}$ is a (sub-)Gaussian matrix. Under strong conditions \cite{vandeGeer.2014} derived results for deterministic and bounded random measurement matrices. Later, the work \cite{ journal2022} improved the sufficient conditions for the debiased estimator for structured random matrices \footnote{A preliminary version of our work was published at the ICASSP 2023 \cite{10096320}.}. The construction of the \emph{debiased LASSO} is based on the KKT conditions and explained in detail in \cite{vandeGeer.2014}. This debiased estimator $\hat{x}$ is defined as
\begin{equation}\label{eq:debiased_lasso}
    \hat{x}^u= \hat{x}+\frac{1}{m}MA^T(b-A\hat{x}),
\end{equation}
where $M$ is a matrix that is chosen in order to ``approximately invert'' the sample covariance matrix $\hat{\Sigma}:=A^TA/m$, i.e. $M \hat{\Sigma} \approx I_{N\times N}$. The main observation is the following decomposition of the difference between the debiased LASSO and the ground truth:
\begin{equation}
    \sqrt{m}(\hat{x}^u - x^*) = \frac{M A^T \varepsilon}{\sqrt{m}} - \sqrt{m}(M \hat{\Sigma} - I_{N\times N})(\hat{x} - x^*).
\end{equation}
The remainder or bias term $ R:=(M \hat{\Sigma} - I_{N\times N})(\hat{x} - x^*) $ asymptotically vanishes which has been shown, e.g. by \cite{Javanmard.2014, Javanmard.2018, vandeGeer.2014}. Since $ M A^T \varepsilon /\sqrt{m} \sim\mathcal{N}(0,\sigma^2M\hat{\Sigma}M) $, it is possible to construct confidence intervals for $x^*$ based on $\hat{x}^u$.

\section{Debiased LISTA}\label{sec:gerneral_estimators}
In this paper, we extend the previous approach for the LASSO estimator to LISTA. We assume the dictionary to have i.i.d. rows, uniformly bounded entries, and, for simplicity, a second-moment matrix $\mathbb{E}[a_1a_1^T]=I_{N\times N}$ for all $i\in[N]$. This assumption is fulfilled by bounded orthonormal systems that encompass several important examples such as the subsampled Fourier matrix used in MRI.

Let $x^k$ be a $k$-th iterate of the LISTA algorithm \eqref{eq:lista}. In view of \eqref{eq:debiased_lasso} we define the $k$-th debiased LISTA iterate $x^k_u$ as
\begin{equation}\label{eq:debiased_estimator}
    x^k_u= x^k+\frac{1}{m}A^T(b-Ax^k).
\end{equation}
Note, that because we required $\mathbb{E}[a_1a_1^T]=I_{N\times N}$, we set $M=I_{N\times N}$ in \eqref{eq:debiased_lasso}. We emphasize that the dictionary matrix $A$ is assumed to be explicitly known.

Since the quality of the LISTA estimator itself as well as the debiased LISTA strongly depends on the $\ell_2$ norm of the difference $x^k-x^*$ we state our main result depending on this quantity. In Section \ref{sec:debiased_LISTA} we provide a detailed analysis for $\ell_2$ consistency. Our main result depends on the number of training samples $n$. In order to simplify the notation we omit this dependence in the following theorem, which will be the core of UQ for LISTA.
\begin{theorem}\label{thm:mainresult}
Let $\frac{1}{\sqrt{m}} A\in \mathbb{R}^{m\times N}$ be a normalized random matrix with i.i.d. rows and uniformly bounded entries, i.e., $\vert A_{ij}\vert\leq K$ for a constant $K\geq 1$ and let \eqref{Cmodel} be the underlying model. Furthermore let the second moment matrix be the identity, i.e. $\mathbb{E}[a_1a_1^*]=I_{N\times N}$.
Then, the following decomposition holds
\begin{equation}
    \sqrt{m}(x^k_u-x^*)=W+R,
\end{equation}
where the $k$-th iterated debiased LISTA $x_u^k$ is defined in \eqref{eq:debiased_estimator}, $W\mid A\sim\mathcal{N}(0,\sigma^2\hat{\Sigma})$ with noise level $\sigma$ from model \eqref{Cmodel} and, if $\vert\supp(x^k-x^*)\vert\leq Cs$, then
\begin{align}
    &\mathbb{P}\left(\Vert R\Vert_{\infty}\geq 4 K \sqrt{\log N} \Vert x^k-x^*\Vert_2\right)\label{eq:prob}\\
    &\leq 2N \exp\left(-\frac{1}{1/(2\log N)+ \sqrt{Cs}/(3\sqrt{m\log N})}\right).\label{eq:tail}
\end{align}
\end{theorem}

The quantities \eqref{eq:prob} and \eqref{eq:tail} depend indirectly on the number of training samples $n$. First, $\Vert x^k-x^*\Vert_2$ depends on $n$, since the reconstruction accuracy is determined by the amount of training data. The constant $C$ also depends on $n$, since usually more training data leads to a better support estimation.
The factor $4K\sqrt{\log N}$ is for normalizing the probability. Since the term $\sqrt{Cs}/(3\sqrt{m\log N})$ vanishes quickly, it remains $2N\exp\left(-\frac{1}{1/(2\log N)}\right) = 2N^{-1}$.

\section{Uncertainty quantification: confidence intervals}\label{sec:conf_regions}

For asymptotically normal estimators there is a standard procedure to construct confidence intervals (CI) (cf. \cite[Chapter 6]{wasserman2013all}). In particular, when the remainder term $R$ vanishes, the construction of CIs comes from the Gaussianity of $W$. This is the case when $4 K \sqrt{\log N} \Vert x^k-x^*\Vert_2$ is small, which requires $\Vert x^k-x^*\Vert_2$ to be small. This is either true when the dictionary provides more information, i.e. the number of rows $m$ is large, or the number of training samples $n$ is large. In this setting, our main theoretical result, Theorem \ref{thm:mainresult}, states that conditioned on $A$, the debiased LISTA $x^k_u$ is asymptotically normal, i.e.
\begin{equation}\label{eq:asym_normal}
    \sqrt{m}(x^k_u-x^*)\mid A\sim\mathcal{N}(0,\sigma^2\hat{\Sigma}).
\end{equation}
We assume to have a consistent noise estimator $\hat{\sigma}$ \cite{Javanmard.2014}. The confidence regions with significance level $\alpha\in(0,1)$ for $x_i^*\in\mathbb{R},\,i\in[N]$, estimated by the debiased LISTA are given by
\begin{equation}\label{eq:conf_region}
    J_i(\alpha):=[(x_u^k)_i-\delta(\alpha)_i, (x_u^k)_i+\delta(\alpha)_i].
\end{equation}
The radius is defined as $\delta(\alpha)_i:=\frac{\hat{\sigma}\hat{\Sigma}_{ii}^{1/2}}{\sqrt{m}}\Phi^{-1}(1-\alpha/2)$, where $\Phi^{-1}$ denotes the quantile function of the standard normal distribution. The CIs constructed in this way are asymptotically valid, i.e., $\lim\limits_{m\to\infty}\mathbb{P}\left(x_i^*\in J_i(\alpha)\right)=1-\alpha.$

The works \cite{Javanmard.2018, Cai.2017} show, that the radii of the CI are optimal as they scale with $\frac{1}{\sqrt{m}}$. The CI are derived straightforwardly from the asymptotic normality. For a more detailed description of the construction see, e.g. \cite{Javanmard.2014, journal2022}.

Note that the CI construction described above is with respect to a single component. This roughly means that if we recover $x^*_i$ $L$ different times based on different data $b$, which is again caused by different realizations of the noise $\varepsilon$, then in $1-\alpha$ of the $L$ cases we construct a confidence interval $J_i(\alpha)$ for $x_i^*$ that contains $x_i^*$.

\section{$\ell_2$-consistency of LISTA}\label{sec:debiased_LISTA}

The probability that the remainder term $R$ vanishes depends on the $\ell_2$-consistency of the LISTA estimator, as Theorem \ref{thm:mainresult} states. Therefore, a crucial analysis of such an estimator is necessary. The first work that established theoretical guarantees for LISTA-type estimators, such as the LISTA-CP, is \cite{chen2018theoretical}. There, however, the authors considered an adversarial noise model, whereas for an asymptotically normal debiased estimator it is essential to assume a noise model that follows a statistical distribution since the asymptotic normality is based on the normal distribution of the noise. Their main result for LISTA-CP assumes the signal and the noise to be in the set $\mathcal{X}(B,s_0,\rho)$ which is defined as
\begin{equation}\label{eq:assumption1}
    \{(x^*,\varepsilon):\Vert x^*\Vert_{\infty}\leq B,\Vert x^*\Vert_0\leq s_0,\Vert\varepsilon\Vert_1\leq \rho \}.
\end{equation}
and reads as
\begin{theorem}\cite[Theorem 2]{chen2018theoretical}\label{l2-consistency_LISTA}\label{thm:l2bound}
    Given $\{W^k,\lambda^k\}_{k=0}^{\infty}$ and $x^0=0$, let $\{x^k\}_{k=1}^{\infty}$ be generated by
    \begin{equation}
        x^{k+1}=\mathcal{S}_{\lambda}\left(x^k+(W^k)^T(b-Ax^k)\right)
    \end{equation}
    If $(x^*,\varepsilon)\in \mathcal{X}(B, s_0, \rho)$ and $s_0$ is sufficiently small, then there exists a sequence of parameters $\{W^k,\lambda^k\}$ such that, for all $(x^*,\varepsilon)\in \mathcal{X}(B,s,\rho)$, we have the error bound:
    \begin{equation}\label{eq:l2_error_bound}
        \Vert x^k-x^*\Vert_2\leq sB\exp(-ck)+C\rho,\qquad \forall k=1,2,\dots ,
    \end{equation}
    where the constants $c>0$, $C>0$ only depend on $A$ and $s$.
\end{theorem}
It is important to highlight that the sequence of parameters obtained in Theorem \ref{l2-consistency_LISTA} may not be the one obtained through empirical risk minimization. Therefore, such results contribute to the architecture utilized rather than the validation of the training process itself. Still, this can be used to estimate the theoretical recovery quality. For the purpose of quantifying the remainder term in Theorem \ref{thm:mainresult} we consider this bound and analyze it further.

The bound was derived using adversarial noise. But it can be adapted for statistical noise by replacing the second term $C\rho$ by $\widetilde{C}\theta_0$ where $\widetilde{C}$ is a constant depending on $A$ and $s$, and $\theta_0$ is a parameter that measures the impact of the (statistical) noise on the weight matrices. This impact is needed to adapt Theorem \ref{thm:l2bound}. The following statement quantifies $\theta_0$:
\begin{theorem}
    Let $\varepsilon\sim\mathcal{N}(0,\sigma^2I_{m\times m})$ be a statistical noise vector and $W^k\in\mathbb{R}^{m\times N}$ with $\Vert W^k_j\Vert_2\leq C_W$. Then for $\theta_0:= C_W\sigma\sqrt{6\log N}$
    $$\mathbb{P}(\max\limits_{j\in[N]} \vert\langle\varepsilon,W^k_j\rangle\vert\geq\theta_0)\leq N^{-2}.$$
\end{theorem}
Thus, in the setting of statistical noise $\varepsilon\sim\mathcal{N}(0,\sigma^2I_{m\times m})$ we derive
\begin{equation}\label{eq:first}
    \Vert x^k-x^*\Vert_2\leq sB\exp(-ck)
    +\widetilde{C}C_W\sigma\sqrt{6\log N}
\end{equation}
The term $sB\exp(-ck)$ in \eqref{eq:first} decreases exponentially in the number of iterations. The second term \eqref{eq:first} depends on $\sigma$ and the noise $\varepsilon$ is assumed to be bounded, cf. \eqref{eq:assumption1}. Therefore, $\sigma$ decreases faster than $\log N$ if $m$ increases (in expectation $\sigma$ decreases like $1/m$), and hence the second term becomes small for large $m$. Therefore, also $4 K \sqrt{\log N}\Vert x^k-x^*\Vert_2$ is small and in view of Theorem \ref{thm:mainresult} large numbers $k$ and $m$ lead to a vanishing remainder term with high probability.

\section{Numerical Results}\label{sec:numerics}
In view of Section \ref{sec:debiased_LISTA}, we experimentally validate the performance of the debiased estimator for LISTA-CP. We test our theory with examples generated from different measurement matrices such as Gaussian and Hadamard matrices.

\subsection{Training process}
For training the LISTA-CP, we rely on the same strategy as \cite{chen2018theoretical}. We reproduce some of the details here for the sake of completeness. We adopt a stage-wise training strategy that is optimized with Adam with a learning rate decay \cite{zheng2022hybrid, chen2018theoretical}. Denoting by $\Theta^{\tau}=\{ (W^k, \theta^k)\}_{k=0}^{\tau}$ all weights in all the iterations up to the the $\tau$-th one. Further, each weight is multiplied by a learning multiplier $c(\,\cdot\,)$ which is initialized as $1$. We define the initial learning rate as $\alpha_0=0.0005$ and two decayed learning rate as $\alpha_1=0.2 \alpha_0$, $\alpha_2=0.02 \alpha_0$. The network is then trained layer by layer. For each layer, we initialize $c(W^{\tau}), c(\lambda^{\tau})=1$ and we pre-train $\tau$, $\Theta^{\tau-1}$. Then, we train $(W^{\tau}, \lambda^{\tau})$ with the initial learning rate $\alpha_0$ and fine-tune $\Theta^{\tau}=\Theta^{\tau-1} \cup \{ (W^{\tau}, \lambda^{\tau})\}$ with the learning rates $\alpha_1$ and $\alpha_2$. Finally,  each weight in $\Theta^{\tau}$ with a decay rate $\gamma=0.3$. We calculate the NMSE$=10 \log_{10} \lVert x^k - x^* \rVert^2 / \lVert x^* \rVert^2$ in every step and, if it has not increased for a long time ($4000$ iterations) or after $200.000$ iterations, we proceed to the next training stage. The reason for that is to stabilize the training process with a learning rate decay. The experiments are implemented using Tensorflow $1.12$ on a workstation with AMD EPYC 7F52 16-Core CPU and NVIDIA A100 PCIe 40GB GPU.\\
For the training data, the support $S$ of the sparse vectors $x^0$ consists of indices which are drawn i.i.d. from a Bernoulli distribution with $p=0.1$ for the value $1$. The values $x_i^0$, $i\in S$, are drawn i.i.d. from a standard Gaussian distribution. Hence, the sparsity is around $ 0.1N$. The measurement matrix $A\in\mathbb{R}^{N\times N}$ with $N=1000$ is subsampled by selecting $m\in\{600,800\}$ rows independently and uniformly at random. The data vectors are obtained by $b=Ax+\varepsilon$, where $\varepsilon$ is i.i.d. Gaussian with SNR$=20$ leading to a relative noise level of $\Vert \varepsilon\Vert_2/\Vert y\Vert_2\approx 0.1$.

\subsection{Experimental setup}

For each of our experiments, we choose a vector $x^*$ generated in the same way as the training samples $x^0$ as the sparse ground truth. We apply $A$ and add i.i.d. Gaussian noise with SNR$=20$ to obtain a data vector $b\in\mathbb{R}^m$ with $m\in\{600,800\}$. We calculate for $k=16$ the debiased LISTA via \eqref{eq:debiased_estimator} and the CIs for every component of $x^*$ are computed via \eqref{eq:conf_region}. Throughout the experiments, we assume for simplicity that $\sigma$ is known and set $\alpha=0.05$.

\subsection{Gaussian measurement matrix}
Although our main result is even valid for data that is generated from structured matrices such as those associated to a BOS \cite{journal2022}, we use a column-normalized Gaussian matrix $A\sim\mathcal{N}(0,\frac{1}{\sqrt{m}}I_{N\times N})$ and select $600$ rows.

\begin{figure}
     \centering
     \begin{subfigure}[b]{0.49\columnwidth}
         \centering
         \includegraphics[width=1\textwidth]{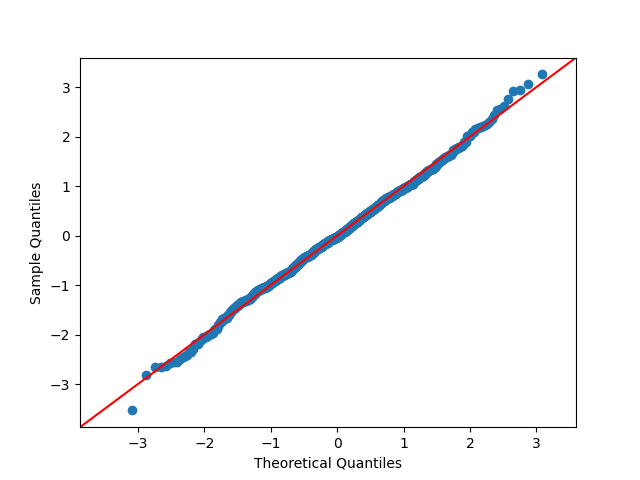}
         \caption{Gaussian matrix}
         \label{fig:QQplot_Gaussian_m600}
     \end{subfigure}
     \begin{subfigure}[b]{0.49\columnwidth}
         \centering
         \includegraphics[width=1\textwidth]{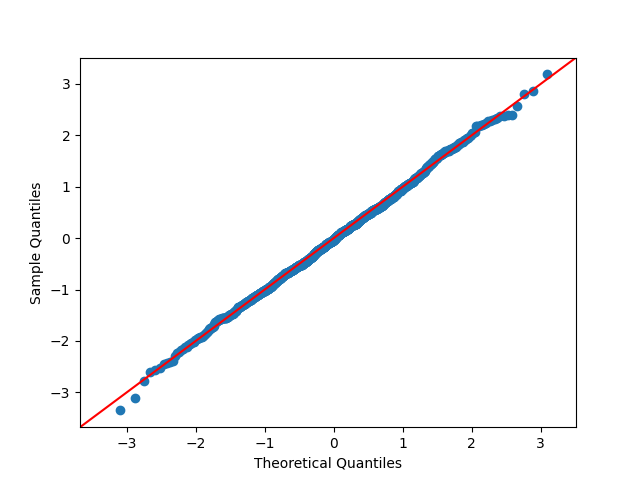}
         \caption{Hadamard matrix}
         \label{fig:QQplot_Hadamard_m800}
     \end{subfigure}
      \caption{Q-Q plot for one realization of $\frac{\sqrt{2m}}{\sigma}(x^{16}_u-x^*)$ vs. the standard normal distribution. The Gaussian matrix has $m=600$ and the Hadamard matrix $800$ rows.}
\end{figure}

In the first experiment, we want to confirm the asymptotic normality as stated in our main result Theorem \ref{thm:mainresult}. For this purpose, we plot the $N$ quantiles of $\frac{\sqrt{2m}}{\sigma}(x^{16}_u-x^*)$, where the randomness here comes from the randomness of the noise, against the $N$ quantiles of the standard normal distribution. The Q-Q plot is illustrated in Figure \ref{fig:QQplot_Gaussian_m600}. The fact that the sorted quantiles lie on the identity line is a strong indicator that both, the quantiles of $\frac{\sqrt{2m}}{\sigma}(x^{16}_u-x^*)$ and the theoretical ones, have the same standard normal distribution.

\begin{figure}
     \centering
     \begin{subfigure}[b]{0.49\columnwidth}
         \centering
         \includegraphics[width=1\textwidth]{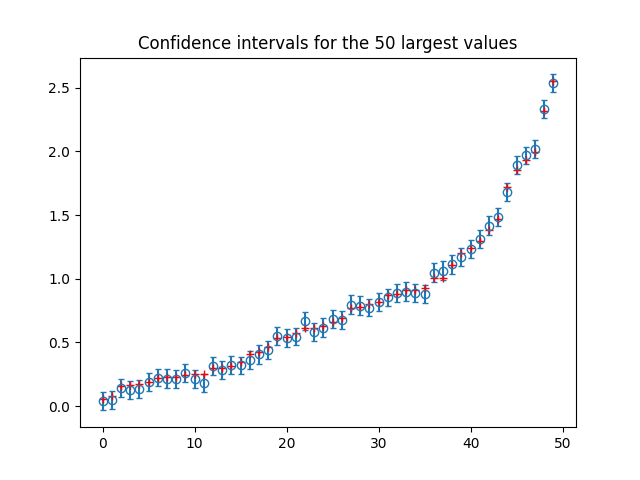}
         \caption{Gaussian matrix}
         \label{fig:conf_intervals_Gaussian}
     \end{subfigure}
     \begin{subfigure}[b]{0.49\columnwidth}
         \centering
         \includegraphics[width=1\textwidth]{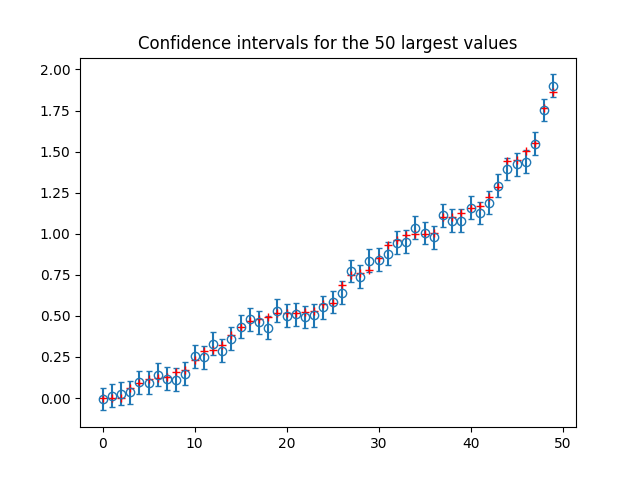}
         \caption{Hadamard matrix}
         \label{fig:conf_intervals_Hadamard}
     \end{subfigure}
      \caption{CIs for the $50$ largest values of one realization of the noise vector. The blue dots are the debiased LISTA for the corresponding components, the red crosses the underlying signal components and the errorbars represent the CIs centered around the debiased LISTA. Both matrices have $m=800$ rows. In the Gaussian case, on the whole support, $99.0\%$ of the CI contain the true parameter. In the Hadamard case the rate is $98.1\%$.}
\end{figure}

In the second step, we check if the CIs are suitable, i.e. if they contain the true parameter with probability $1-\alpha$. Although we discussed in Section \ref{sec:conf_regions} that the statement refers to a single ground truth component, we want to measure how reliable the CIs for the entire ground truth vector $x^*$ based on one noise realization of the data vector $b$ is. Therefore, we define hitrates for all components $h = \frac{1}{N}\sum_{i\in [N]}\mathbbm{1}_{\{x_i^*\in J_i\}}$ and for the components of the support $h_S= \frac{1}{s}\sum_{i\in S}\mathbbm{1}_{\{x_i^*\in J_i\}}$ respectively.
We conduct the recovery and debiasing process $500$ times, each time with a different noise realization and a debiased LISTA with CIs based on this noise realization. Then we calculate the hitrates and average over the number of conducted experiments, i.e., $500$. We obtain $h = 0.984$ and $h_S = 0.900$. Based on the CI construction, we would expect that $h_S$ is around the predicted $0.95$. But this statement is of asymptotic nature. In order to increase the hitrates we could select more rows of the subsampled measurement matrix, i.e., increase $m$, provide more training samples, i.e., increase $n$ or increase the number of iterates. For example, if we set $m=800$ we obtain hitrates $h=0.999$ and $h_S=0.996$ (cf. Table \ref{tab:hitrates}). Figure \ref{fig:conf_intervals_Gaussian} presents for one realization of the noise the 50 largest ground truth components with the corresponding debiased LISTA components and CIs. 

\vspace{-.2cm}
\subsection{Hadamard measurement matrix}
The data acquisition process of a magnetic resonance scanner is modelled via a subsampled Fourier measurement matrix \cite{lustig2008compressed}. In order to simplify the discussion due to the lack of space, in this paper we run experiments for real measurements and, therefore, we use Hadamard matrices, which can be interpreted as a Fourier transform on $\{0,1\}^d$ \cite[Chapter 12]{Foucart.2013}. It is defined recursively, for $l = 2^d$, as
\begin{equation}\label{eq:hadamard_matrix}
    H_d = \begin{pmatrix} H_{d-1} & H_{d-1} \\ H_{d-1} & -H_{d-1}\end{pmatrix},\qquad H_0=1,
\end{equation}
and satisfies the requirements of Theorem \ref{thm:mainresult}. We conduct the same experiment as above with 500 different noise realizations. The averaged hitrates are shown in Table \ref{tab:hitrates}. The observation that the debiased LISTA for signals acquired with a Hadamard matrix performs worse than in the Gaussian case is not surprising since it is harder for random structured matrices to fulfill the sufficient conditions required for signal recovery. In order to increase the hitrates we run the same experiments with more measurements ($m=800$). Figure \ref{fig:QQplot_Hadamard_m800} and \ref{fig:conf_intervals_Hadamard} show the Q-Q plot and CIs in the Hadamard case.

\begin{table}[t]
\small
    \centering
    \begin{tabular}{c|c|c|c|c}
         Measurement & $m$ & $h_S$ & $h$ \\
         \hline 
         Gaussian & 600 & 0.900 & 0.984 \\
         Gaussian & 800 & 0.996 & 0.999 \\
         Hadamard & 600 & 0.843 & 0.975 \\
         Hadamard & 800 & 0.909 & 0.985
    \end{tabular}
    \caption{Values of $h_{S_0}$, $h$ for measurement matrices with different numbers of rows and training data. The values are averaged over 500 realizations of the noise with SNR $=20$.}
    \label{tab:hitrates}
\end{table}

\section{Conclusion}
In this paper, we contribute to the field of interpretability in AI by developing a rigorous framework for how to de-bias the learned ISTA estimator. We show that the debiased LISTA is asymptotically Gaussian distributed which allows for uncertainty quantification, more precisely for the construction of CIs. We confirm our theoretical results with numerical experiments using Gaussian and Hadamard measurement matrices. This contributes, from the signal processing and statistical point of view to the understanding of limitations and potential failures of neural networks that will be used in critical applications.

\bibliographystyle{IEEEbib}
\bibliography{strings,refs}

\end{document}